\documentclass[letterpaper]{article}
\usepackage{aaai2027}

\usepackage[hyphens]{url}
\usepackage{graphicx}
\usepackage{natbib}
\usepackage{caption}
\usepackage{amsmath}
\usepackage{amssymb}
\usepackage{booktabs}
\usepackage{array}

\newcommand{\method}{OSAGEN}
\newcommand{\QBG}{Query-Bias-Gen}
\newcommand{\ISC}{\textsc{ISC}}
\newcommand{\Enc}{\operatorname{Enc}}

\newcommand{\Core}{\operatorname{Core}}
\newcommand{\Outer}{\operatorname{Outer}}

\title{OSAGEN: Object-Aware Mask Priors and Multistage Decoupled Diffusion for Industrial Anomaly Generation}
\author{Anonymous Submission}
\affiliations{}

\author{
    Jinyi Xu\textsuperscript{1},
    Peng Chen\textsuperscript{1},
    Yunkang Cao\textsuperscript{2},
    Chengliang Liu\textsuperscript{3},\\
    Xinghui Dong\textsuperscript{4},
    Chao Huang\textsuperscript{1}\thanks{Corresponding author. Email: huangch253@mail.sysu.edu.cn}
}

\affiliations{
    \textsuperscript{1}School of Cyber Science and Technology, Shenzhen Campus of Sun Yat-sen University\\
    \textsuperscript{2}School of Artificial Intelligence and Robotics, Hunan University\\
    \textsuperscript{3}Department of Computer and Information Science, University of Macau\\
    \textsuperscript{4}School of Computer Science and Technology, Ocean University of China
}

\begin{document}

\maketitle

\begin{abstract} Industrial anomaly detection and localization are limited by scarce real anomalies and pixel-level annotations, a bottleneck that synthetic image--mask pairs can alleviate. However, existing few-shot mask-guided generation may over-follow mask geometry, produce weak anomalies, or use condition masks incompatible with the current object instance. We propose \method{}, which combines object-aware mask priors with multistage decoupled diffusion. Its three-stage adaptation sequentially learns normal appearance, defect appearance under coarse conditions, and fine-grained mask calibration, improving defect realization and local control. \QBG{} injects object structure from a matched normal image into mask diffusion to produce object-aware priors, while \ISC{} restricts anomaly propagation and preserves normal content during sampling. A lightweight materialization step recovers pixel-level labels aligned with the realized defects. On MVTec AD and VisA, \method{} achieves AP-P/F1-P scores of 88.1/82.2 and 68.5/66.1, respectively, under a unified downstream localization protocol. The code will be released upon acceptance. 
\end{abstract}

\section{Introduction}

Industrial visual inspection requires reliable image-level detection and
pixel-level localization, but real anomalies and their masks are scarce
in practical production settings~\cite{jin2026reasoning,lu2026maskad,
wang2026normal,chen2026dyc}. Synthetic image--mask pairs can alleviate
this bottleneck by providing supervised data for downstream inspection
training. Beyond producing visually plausible images, localized anomaly
generation must preserve normal object content, realize clear defect
semantics, and provide spatially meaningful pixel-level supervision.

\begin{figure*}[t]
\centering
\includegraphics[width=0.95\textwidth]{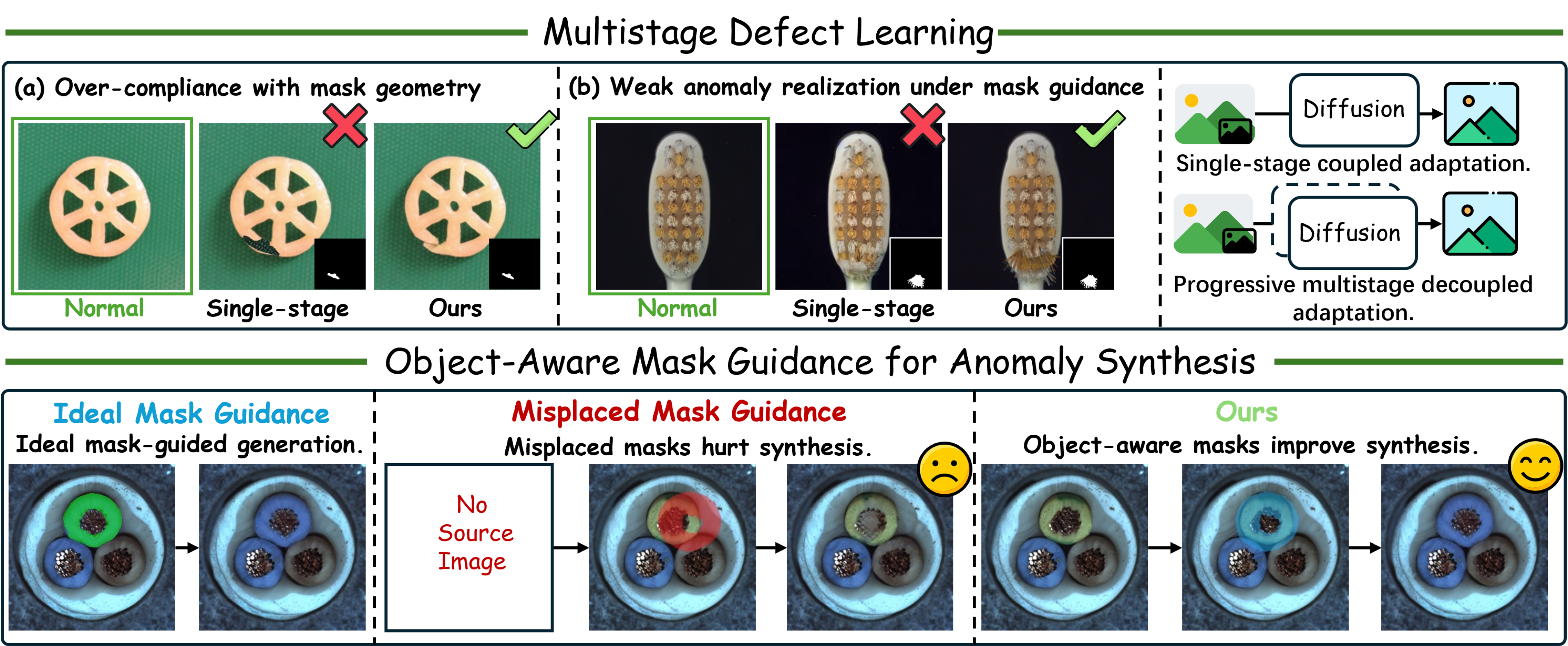}
\caption{
Motivation of \method{}. Top: single-stage adaptation jointly
learns normal appearance, defect semantics, and mask compliance. Because
inference-time condition masks vary in shape and extent and need not
coincide with the visible anomaly boundary, coupling defect
generation to their exact geometry may produce mask-shaped, weak,
incomplete, or failed defects. \method{} instead separates normal
learning, defect learning under a coarse condition, and fine-grained
mask calibration through multistage learning. Bottom: a mask
generated without the context of the current object image may be
misplaced and impair anomaly synthesis, whereas \QBG{}
provides object-aware guidance for the target image.
}
\label{fig:motivation}
\end{figure*}

Recent diffusion-based methods have advanced few-shot
industrial anomaly generation through personalized adaptation,
normal--anomaly separation, mask-guided inpainting, joint image--mask
modeling, and inference-time guidance~\cite{hu2023anomalydiffusion,
dai2025SeaS,song2025defectfill,choi2026magic,rao2026one}. Nevertheless,
two issues remain insufficiently addressed in mask-guided few-shot
generation. First, single-stage adaptation learns normal
appearance, defect semantics, and mask compliance together, tightly
coupling defect learning to the geometry of the condition mask.
Inference-time condition masks may vary in shape and extent and are
intended to guide the editable region rather than prescribe the exact
visible anomaly boundary. Therefore, forcing the generated defect to
strictly follow their geometry can be counterproductive: the model may
mechanically inherit the mask shape, realize only a weak or incomplete
anomaly, or fail to generate the defect altogether. Second, in many
mask-guided pipelines, condition masks are externally supplied
or sampled from defect-category-conditioned generators, while mask
generation is not directly conditioned on the current object image. As
illustrated in Figure~\ref{fig:motivation}, the resulting mask is usually misplaced or otherwise incompatible with anomaly synthesis.

We address both issues with \method{}. Its three-stage curriculum learns
normal appearance, defect semantics under a coarse condition, and
fine-grained mask calibration, thereby reducing dependence on exact
condition geometry. \QBG{} injects structure from a matched normal image
into mask-diffusion queries to produce object-aware priors. During inference, \ISC{} reinstates the training-time spatial constraints
without supervision, keeping anomaly semantics inside the prior and
preserving surrounding content, while a materialization step converts the
loose condition into a pixel-level label aligned with the visible anomaly
without altering the generated image.
Our contributions are:
\begin{itemize}
    \item We design multistage decoupled defect-conditioned
    inpainting that sequentially separates normal learning, defect
    learning under a coarse condition, and fine-grained mask
    calibration. The resulting branches and spatial constraints further support \ISC{}
    for localized anomaly realization during inference.

    \item We propose \QBG{}, which injects object structure from a
    structurally matched normal image into mask diffusion to produce
    object-aware mask priors conditioned on the current object instance.

    \item \method{} constructs effective synthetic image--mask pairs
    and achieves strong downstream anomaly localization performance on
    MVTec AD and VisA.
\end{itemize}

\section{Related Work}

\textbf{Industrial anomaly inspection and synthetic supervision.}
Synthetic anomaly generation provides a complementary way to construct
supervised image--mask pairs for downstream
inspection~\cite{shi2026omni,li2026gpflow,qian2026quality}. Early model-free and GAN-based
synthesis spans patch or texture perturbation, feature-space synthesis,
and learned defect translation~\cite{cutpaste,draem,nsa,glass,realnet,
dfmgan,sdgan,defectgan}. These methods demonstrate the usefulness of
synthetic supervision, while jointly achieving object-compatible mask
placement, realistic defect appearance, and accurate pixel-level labels
remains challenging.

\textbf{Diffusion anomaly generation.}
Diffusion backbones and personalization techniques, including latent
diffusion, Textual Inversion, DreamBooth, and
LoRA~\cite{latentdiffusion,textualinversion,dreambooth,lora}, have
become strong foundations for few-shot anomaly
synthesis~\cite{lu2025generate,shin2025anodapter}. Recent industrial
generators span anomaly-embedding or mask-conditioned adaptation, joint
image--mask or whole-image generation, local inpainting, and
training-free control~\cite{hu2023anomalydiffusion,ali2024anomalycontrol,
anomalyxfusion,anogen,dualanodiff,dai2025SeaS,song2025defectfill,
choi2026magic,rao2026one,anomalyany,tf2}. Mask-guided approaches provide
explicit local control and commonly obtain condition masks from supplied
annotations or defect-category-conditioned generators, without
conditioning on the current object image, and typically couple normal
appearance, defect semantics, and mask compliance within a single
adaptation stage. In contrast, \method{} combines object-aware mask priors with multistage decoupled inpainting, while \ISC{} exploits the normal and defect branches to restrict inference-time anomaly propagation.

\section{Method}

\begin{figure*}[t]
\centering
\includegraphics[width=0.98\textwidth]{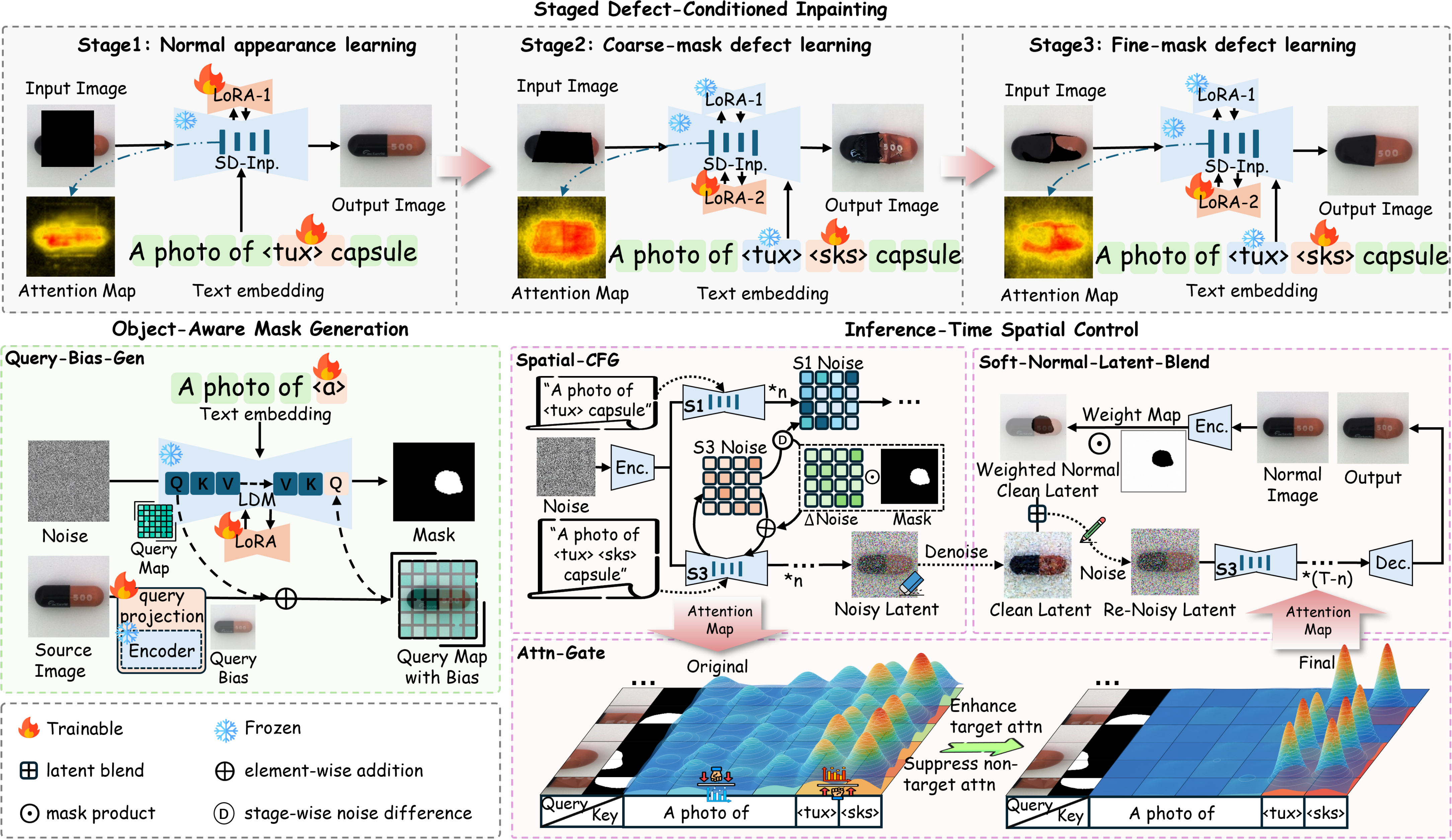}
\caption{\method{} overall framework with three components. Staged inpainting (top) learns normal appearance,
defect semantics under coarse support, and fine-grained mask calibration. \QBG{} (bottom left) injects source-image structure into mask diffusion to produce an object-aware mask prior. Inference-time Spatial Control (\ISC{}) (bottom right): Spatial-CFG amplifies defects inside the support via the S3-minus-S1 noise-prediction residual; Attn-Gate strengthens target-token attention and suppresses irrelevant tokens; Soft-Normal-Latent-Blend pulls the background latent toward the source image.}
\label{fig:overview}
\end{figure*}

\subsection{Problem Formulation}

Let the training set contain normal images $\mathcal{X}_n$ and few
anomalous image--mask pairs
$\mathcal{D}_{\mathrm{anom}}=\{(x_a,m_{gt})\}$
from object category $o$ and defect category $a$. We generate synthetic
pairs $\{(\tilde{x}_a,\tilde{y}_a)\}$ for downstream inspection without
using test anomalies or masks. The condition mask is a spatial prior,
not the final label; materialization recovers supervision aligned with
the visible defect. The main generator $G_{\theta}$ is based on Stable
Diffusion v1.5 inpainting~\cite{latentdiffusion}: given a clean latent
$x_0$, the forward process samples
\begin{equation}
    x_t=\sqrt{\bar{\alpha}_t}\,x_0+\sqrt{1-\bar{\alpha}_t}\,\epsilon,\qquad
    \epsilon\sim\mathcal{N}(0,I),
    \label{eq:forward_diffusion}
\end{equation}
where $t$ is the diffusion timestep and $\bar{\alpha}_t$ its
noise-schedule coefficient; clean-latent estimates below use one-step
recovery of $x_0$ from $x_t$ and the predicted noise.

\subsection{Staged Defect-Conditioned Inpainting}

A single mask-guided stage entangles normal appearance, defect semantics,
and mask response. SeaS separates normal-product patterns and anomaly
attributes within a shared U-Net~\cite{dai2025SeaS}; in our setting,
the three-stage curriculum assigns normal appearance to
\texttt{<tux>} and LoRA-1 and defect semantics to \texttt{<sks>} and
LoRA-2 (Figure~\ref{fig:overview}; Table~\ref{tab:stage_setup}). The staged curriculum progressively reduces objective interference across stages and weakens the dependence of defect learning on exact condition geometry throughout adaptation.

\begin{table}[t]
\centering
\small
\setlength{\tabcolsep}{2.2pt}
\begin{tabular}{lccccc}
\toprule
$s$ & $x^s$ & $m_c^s$ & $m_l^s$ & Prompt tokens & Trainable \\
\midrule
1 & $x_n$ & $m_r$ & $m_r$ & \texttt{<tux>, obj} & \texttt{<tux>}, LoRA-1 \\
2 & $x_a$ & $m_{\mathrm{coarse}}$ & $m_{gt}$ & \texttt{<tux>, <sks>, obj} & \texttt{<sks>}, LoRA-2 \\
3 & $x_a$ & $m_{gt}$ & $m_{gt}$ & \texttt{<tux>, <sks>, obj} & \texttt{<sks>}, LoRA-2 \\
\bottomrule
\end{tabular}
\caption{Three-stage inpainting setup. Prompts use the fixed prefix ``A photo of''; the table lists the variable prompt tokens. $m_r$ is a random mask, $m_{\mathrm{coarse}}$ is a coarse condition region, and $m_{gt}$ is the ground-truth anomaly mask. \texttt{<tux>} binds object appearance, while \texttt{<sks>} binds defect semantics.}
\label{tab:stage_setup}
\end{table}

\begin{figure}[t] 
\centering \includegraphics[width=0.95\columnwidth]{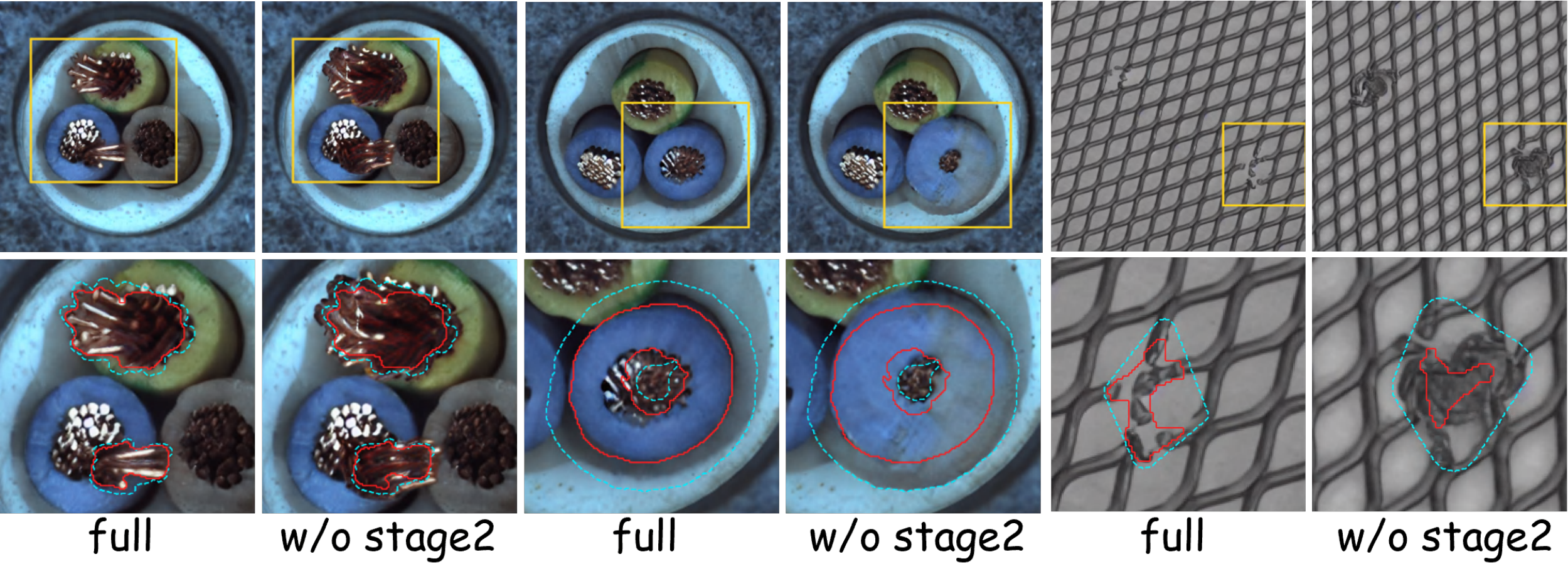} \caption{Visual analysis of the staged inpainting curriculum. Without Stage~2, the model tends to spread defects over the coarse region or introduce artifacts, whereas the full model confines the visible anomaly to a more plausible subregion.} \label{fig:stage2_visual} \end{figure}

Stage~1 learns normal appearance from normal images $x_n$ and random
masks $m_r$, whose varying extent prevents the normal branch
from tying a particular condition geometry to a particular appearance.
Only the object token \texttt{<tux>} and LoRA-1 are updated, producing a
normal branch that anchors defect learning and serves as the
normal-prediction baseline for Spatial-CFG in \ISC{}. 
Stage~2 learns
defect semantics while decoupling the condition-mask extent from the true
anomaly extent. The coarse condition mask $m_{\mathrm{coarse}}$ defines a
loose candidate region, whereas the denoising loss and attention
constraint are applied only inside $m_{gt}$. Consequently, \texttt{<sks>}
and LoRA-2 bind to anomalous appearance rather than the normal content in
$m_{\mathrm{coarse}}\setminus m_{gt}$. LoRA-1 is frozen, and only
\texttt{<sks>} and LoRA-2 are updated. As illustrated in
Figure~\ref{fig:stage2_visual}, this stage teaches the model to treat the
coarse condition mask as a candidate region rather than as the boundary
of the defect. The grouped-minquad construction, its automatic
feasibility fallback, and the Stage~2 preprocessing procedure are detailed in Appendix~D of the supplementary material, together with the corresponding pseudocode. 
Stage~3 sets both
the condition mask and the loss mask to $m_{gt}$, adapting the defect
branch to fine-grained mask conditions. During the final adaptation stage, fine-grained masks reduce the editing burden of overly large conditions and narrow the gap between the mask prior and the visible anomaly, thereby aligning the model with the inference-time mask distribution and facilitating reliable pixel-level labels at materialization.

At stage $s\in\{1,2,3\}$, each training sample is represented as $(x^s,m_c^s,m_l^s,y^s)$, where $x^s$ is the target image, $m_c^s$ is the condition mask specifying the inpainting region, $m_l^s$ is the loss mask specifying where defect supervision is applied, and $y^s$ is the text prompt. The clean latent $x_0^s=\Enc(x^s)$ is noised into $x_t^s$, and the SD-inpainting U-Net predicts the noise from the concatenation of the noisy latent, downsampled condition mask, and masked-image latent:
\begin{equation}
    \hat{\epsilon}^s =
    \epsilon_{\theta}^s
    \left(
    [x_t^s,\mathrm{Down}(m_c^s),
    \Enc(x^s\odot(1-m_c^s))],
    y^s,t
    \right).
    \label{eq:inpaint_input}
\end{equation}
All stages use the same objective:
\begin{equation}
    \mathcal{L}_{\mathrm{stage}}^s
    =\alpha_s\mathcal{L}_{\epsilon}^s
    +\beta_s\mathcal{L}_{\mathrm{bg}}^s
    +\gamma_s\mathcal{L}_{\mathrm{attn}}^s ,
    \label{eq:stage_loss}
\end{equation}
where $\mathcal{L}_{\epsilon}$ learns local appearance inside the loss
region, $\mathcal{L}_{\mathrm{bg}}$ suppresses background drift outside
the condition region, and $\mathcal{L}_{\mathrm{attn}}$ prevents
non-target text tokens from explaining the anomalous region. Before
evaluating each term, the masks are resized to the spatial resolution
of the corresponding latent or attention map; for brevity, we retain
the notation $m_c^s$ and $m_l^s$ after resizing:
\begin{equation}
\begin{aligned}
\mathcal{L}_{\epsilon}^s
&=
\mathbb{E}_{t,\epsilon}
\left[
\frac{\|m_l^s\odot(\hat{\epsilon}^s-\epsilon)\|_2^2}
{|m_l^s|+\xi}
\right],\\
\mathcal{L}_{\mathrm{bg}}^s
&=
\mathbb{E}_{t,\epsilon}
\left[
\frac{\|(1-m_c^s)\odot(\hat{x}_0^s-x_0^s)\|_2^2}
{|1-m_c^s|+\xi}
\right],\\
\mathcal{L}_{\mathrm{attn}}^s
&=
\mathbb{E}_{t}
\left[
\frac{1}{|m_l^s|+\xi}
\sum_{i\in m_l^s}
\sum_{j\in T_{\mathrm{nt}}^s} A_{ij}
\right].
\end{aligned}
\label{eq:stage_terms}
\end{equation}
Here, $\hat{x}_0^s$ is recovered from $\hat{\epsilon}^s$ using the one-step denoising estimate, $A_{ij}$ is the cross-attention weight from spatial position $i$ to token $j$, $T_{\mathrm{nt}}^s$ denotes the non-target token set, and $\xi$ avoids division by empty masks. The staged generator thus learns to realize defects under coarse-to-fine mask conditions, but inference still requires a mask prior compatible with the specific source instance and a plausible defect region. We address this requirement with \QBG{}.

\subsection{Object-aware Mask Generation: Query-Bias-Gen}

The staged inpainting generator requires an inference-time condition mask compatible with the current object instance. Learning a defect-category-conditioned distribution $p(m \mid a)$ from real anomaly masks captures defect-area, connectivity, and shape statistics, but does not ensure compatibility with the structure of a specific normal source image. We therefore formulate mask generation as an object-aware conditional distribution $p(m \mid x_n,a)$, where $x_n$ is the normal source image and $a$ is the defect category. This converts an object-agnostic defect-shape prior into a mask prior conditioned on the current object instance.

We implement this idea with \QBG{} on top of an AnomalyDiffusion-style mask latent diffusion model. As shown in Figure~\ref{fig:overview} (bottom left), the source image $x_n$ is encoded by a frozen visual encoder and passed through a lightweight trainable mapping layer to align its object-structure representation with the cross-attention query tensors of the mask LDM. The aligned source structure is then injected as an additive query bias:
\begin{equation}
    q_l' = q_l + \Delta q_l(x_n),
    \label{eq:qbg_query_bias}
\end{equation}
where $q_l$ is the original query of the $l$-th attention block and $\Delta q_l(x_n)$ denotes the layer-wise query bias computed from the source image. For compactness, $\Delta q(x_n)$ denotes the collection of query biases $\{\Delta q_l(x_n)\}_l$ injected into the selected attention blocks. This keeps the backbone unchanged while biasing spatial tokens with source-object context before they interact with the defect condition.
During training, each real mask $m_{gt}$ is paired with a structurally matched normal source image $x_n$, and the mask diffusion model is trained with the standard denoising objective:
\begin{equation}
    \mathcal{L}_{Q} =
    \mathbb{E}_{t,\epsilon}\big[
    \|\epsilon_{\phi}(z_t^m,t,a,\Delta q(x_n))-\epsilon\|_2^2
    \big],
    \label{eq:qbg_loss}
\end{equation}
where $z_t^m$ is the noisy mask latent of $m_{gt}$ and $\epsilon_{\phi}$ denotes the mask diffusion network with the query-bias injection in Eq.~\eqref{eq:qbg_query_bias}. Its trainable parameters include the mapping layer and the LoRA modules inserted into the mask diffusion network, while the remaining backbone parameters are frozen. 
Given a sampled normal source image $x_n$, defect category $a$, and random noise $\eta$, \QBG{} samples the object-aware mask prior as
\begin{equation}
    m = D_{\mathrm{mask}}\!\left(a,\,\Delta q(x_n),\,\eta\right).
    \label{eq:qbg_infer}
\end{equation}
At inference, this object-aware mask is used as the spatial support for the staged inpainting generator described above.

\subsection{\ISC{}: Inference-Time Spatial Control}

The staged curriculum produces normal and defect branches together
with two training-time spatial constraints. At inference, these
constraints are no longer available as supervision, while the \QBG{}
mask prior specifies only where a defect may occur, not how anomaly
semantics propagate during denoising or how well content outside the
support is preserved. \ISC{} therefore transfers the learned structure
to three sampling-time operations. Attn-Gate acts throughout denoising
as the inference-time counterpart of
$\mathcal{L}_{\mathrm{attn}}$, modulating object and defect tokens that
have already been regularized during adaptation. Spatial-CFG contrasts
the Stage~3 defect branch with the Stage~1 normal branch at timesteps
$\mathcal{T}_{\mathrm{cfg}}$, while Soft-Normal-Latent-Blend transfers
the background-preservation role of $\mathcal{L}_{\mathrm{bg}}$ to
timesteps $\mathcal{T}_{\mathrm{blend}}$. 

\textbf{Attn-Gate.}
Let $A_{ij}^R$ denote the cross-attention weight from position $i$ to token $j$ at attention resolution $R$, and let $T_{\mathrm{target}}$ be the target-token set of learned object and defect tokens. Given mask prior $m$, we project it to the attention grid and obtain a core region $\Core_R(m)$ and a protected region $\Outer_R(m)$, where $\Core_R(m)\subseteq\Outer_R(m)$. We define a spatial gate $g_i^R$ and re-normalize the target-token attention:
\begin{equation}
\begin{aligned}
g_i^R &=
\begin{cases}
\alpha_{\mathrm{in}}^R,  & i\in\Core_R(m),\\
\alpha_{\mathrm{out}}^R, & i\notin\Outer_R(m),\\
1,                       & \mathrm{otherwise},
\end{cases}
\widetilde{A}_{ij}^R
=
\frac{\rho_{ij}^R A_{ij}^R}
{\sum_k \rho_{ik}^R A_{ik}^R}.
\end{aligned}
\end{equation}
Here, $\alpha_{\mathrm{in}}^R>1$, $\alpha_{\mathrm{out}}^R<1$, and the gate is applied only to target tokens: $\rho_{ij}^R=g_i^R$ for $j\in T_{\mathrm{target}}$, and $\rho_{ij}^R=1$ otherwise. This strengthens object and defect tokens inside the anomaly core while suppressing their leakage into the far background.

\textbf{Anomaly-Region Spatial-CFG.}
Spatial-CFG enhances defect semantics at the noise-prediction level. At each timestep $t\in\mathcal{T}_{\mathrm{cfg}}$, the Stage~3 defect and Stage~1 normal branches produce noise predictions $\epsilon_t^{\mathrm{defect}}$ and $\epsilon_t^{\mathrm{normal}}$, respectively. Rather than an unconditional prediction, which describes generation
without the defect prompt rather than a normal instance of the current
object, we take $\epsilon_t^{\mathrm{normal}}$ as the baseline and treat
the branch difference as a residual:
\begin{equation}
\begin{aligned}
\hat{\epsilon}_t
&=
\epsilon_t^{\mathrm{normal}}
+\omega_t M_{\mathrm{ano}}(m)\odot
\left(
\epsilon_t^{\mathrm{defect}}
-\epsilon_t^{\mathrm{normal}}
\right).
\end{aligned}
\label{eq:spatial_cfg}
\end{equation}
Here, $\hat{\epsilon}_t$ is the guided noise prediction used for the denoising update and subsequent clean-latent recovery, $M_{\mathrm{ano}}(m)$ is the soft anomaly-support map, and $\omega_t$ is the time-dependent guidance strength. Since the residual is injected only inside $M_{\mathrm{ano}}(m)$, Spatial-CFG amplifies defect appearance without globally changing the normal object structure.

\textbf{Background Soft-Normal-Latent-Blend.}
$\mathcal{L}_{\mathrm{bg}}$ constrains drift on the predicted clean
latent; we transfer this choice to inference by blending clean latents
rather than overwriting pixels or mixing noisy latents. At each timestep $t\in\mathcal{T}_{\mathrm{blend}}$, we recover the clean-latent estimate $\hat{x}_{0,t}$ from the current latent $x_t$ and the guided noise prediction $\hat{\epsilon}_t$ by rearranging Eq.~\eqref{eq:forward_diffusion}. We then blend the background clean content toward the source-image latent:
\begin{equation}
\begin{aligned}
w_t
&=\lambda_t M_{\mathrm{bg}}(m),\\
\tilde{x}_{0,t}
&=(1-w_t)\hat{x}_{0,t}
+w_t x_0^{\mathrm{src}}.
\end{aligned}
\label{eq:soft_blend_clean}
\end{equation}
Here, $M_{\mathrm{bg}}(m)$ is the soft background-preservation map, $x_0^{\mathrm{src}}$ is the latent of the normal source image, and $\lambda_t$ is the blending strength. The blended clean latent is then re-noised to the current timestep using the same guided noise prediction:
\begin{equation}
\begin{aligned}
x_t \leftarrow{}&
\sqrt{\bar{\alpha}_t}\,
\tilde{x}_{0,t}
+
\sqrt{1-\bar{\alpha}_t}\,
\hat{\epsilon}_t .
\end{aligned}
\label{eq:soft_blend_renoise}
\end{equation}
Fixed attention resolutions, timestep schedules, spatial-weight
construction, and dataset-level guidance strengths are given in
Appendix~G; Appendix~H reports sensitivity to the joint control strength.

\subsection{Label Materialization}
Before inference, the mask sampled by \QBG{} is further morphologically
augmented to form a loose condition region. Together with the
candidate-region formulation of Stage~2, this makes the condition mask
intentionally broader and morphologically less precise than the realized
defect. It therefore cannot serve as the final pixel-level label. We
instead materialize the label from the source-to-generation difference,
encoded both in pixel space and by a frozen DINOv2 backbone, together
with the condition mask. A lightweight predictor outputs
anomaly probabilities inside the condition region, which are thresholded
into the label; the generated image is left unchanged. The per-object predictor is trained for only a few minutes on the
permitted training split, using real anomalies and degraded masks that
simulate generation-time conditions. It draws only on the same permitted
anomalies available to every method and adds no external data.
Implementation details are provided in Appendix~K.

\begin{table*}[t]
\centering
\small
\begin{tabular}{@{}l*{5}{@{\hspace{4pt}}c@{\hspace{1.6pt}}c@{\hspace{1.6pt}}c@{\hspace{1.6pt}}c}@{}}
\toprule
 & \multicolumn{4}{c}{\shortstack{AnomalyDiffusion\\(AAAI 2024)}} & \multicolumn{4}{c}{\shortstack{DualAnoDiff\\(CVPR 2025)}} & \multicolumn{4}{c}{\shortstack{SeaS\\(ICCV 2025)}} & \multicolumn{4}{c}{\shortstack{O2MAG\\(CVPR 2026)}} & \multicolumn{4}{c}{\shortstack{\method{}\\(Ours)}} \\
\cmidrule(lr){2-5}\cmidrule(lr){6-9}\cmidrule(lr){10-13}\cmidrule(lr){14-17}\cmidrule(lr){18-21}
Category & AP-I & AUC-P & AP-P & F1-P & AP-I & AUC-P & AP-P & F1-P & AP-I & AUC-P & AP-P & F1-P & AP-I & AUC-P & AP-P & F1-P & AP-I & AUC-P & AP-P & F1-P \\
\midrule
bottle & 99.9 & 99.4 & 94.1 & 87.3 & 100.0 & 99.5 & 93.4 & 85.7 & 99.9 & 99.7 & 95.9 & 88.8 & 100.0 & 99.7 & 95.4 & 88.4 & 100.0 & 99.8 & 97.0 & 90.9 \\
cable & 100.0 & 99.2 & 90.8 & 83.5 & 98.3 & 98.5 & 82.6 & 76.9 & 98.8 & 96.0 & 83.1 & 77.7 & 100.0 & 99.4 & 91.2 & 85.0 & 99.7 & 98.1 & 87.0 & 81.5 \\
capsule & 99.9 & 98.8 & 57.2 & 59.8 & 99.2 & 99.5 & 73.2 & 67.0 & 99.2 & 93.7 & 41.9 & 47.3 & 99.8 & 97.0 & 60.6 & 59.0 & 99.7 & 98.3 & 66.6 & 62.3 \\
carpet & 98.8 & 98.6 & 81.2 & 74.6 & 99.9 & 99.4 & 89.1 & 80.2 & 99.0 & 99.3 & 86.4 & 78.1 & 100.0 & 99.5 & 88.5 & 80.0 & 99.7 & 99.4 & 88.6 & 80.7 \\
grid & 99.5 & 98.3 & 52.9 & 54.6 & 99.7 & 98.5 & 57.2 & 54.9 & 99.9 & 99.7 & 76.3 & 70.0 & 100.0 & 99.6 & 78.6 & 71.6 & 100.0 & 99.2 & 73.7 & 69.0 \\
hazelnut & 99.9 & 99.8 & 96.5 & 90.6 & 100.0 & 99.8 & 97.7 & 92.8 & 99.8 & 99.5 & 92.3 & 85.6 & 100.0 & 99.8 & 96.2 & 90.1 & 100.0 & 99.8 & 96.7 & 92.1 \\
leather & 100.0 & 99.8 & 79.6 & 71.0 & 100.0 & 99.9 & 88.8 & 78.8 & 100.0 & 99.8 & 85.2 & 77.0 & 100.0 & 99.7 & 88.0 & 79.7 & 100.0 & 99.9 & 90.8 & 83.1 \\
metal\_nut & 100.0 & 99.8 & 98.7 & 94.0 & 99.9 & 99.6 & 98.0 & 93.0 & 100.0 & 99.8 & 99.2 & 95.7 & 100.0 & 99.8 & 99.2 & 95.7 & 100.0 & 99.8 & 99.0 & 95.9 \\
pill & 99.6 & 99.7 & 93.9 & 90.8 & 99.0 & 99.6 & 95.8 & 89.2 & 99.6 & 99.9 & 97.1 & 90.7 & 99.7 & 99.7 & 96.1 & 89.9 & 99.9 & 99.9 & 97.7 & 92.3 \\
screw & 97.9 & 97.0 & 51.8 & 50.9 & 95.0 & 98.1 & 57.1 & 56.1 & 98.0 & 98.5 & 58.5 & 57.2 & 98.2 & 99.4 & 68.2 & 64.4 & 97.6 & 98.2 & 76.9 & 71.6 \\
tile & 100.0 & 99.2 & 93.6 & 86.2 & 100.0 & 99.7 & 97.1 & 91.0 & 100.0 & 99.8 & 97.9 & 92.5 & 100.0 & 99.9 & 98.2 & 92.7 & 99.9 & 99.5 & 95.7 & 89.0 \\
toothbrush & 100.0 & 99.2 & 76.5 & 73.4 & 99.7 & 98.2 & 68.3 & 68.6 & 100.0 & 98.4 & 70.0 & 68.1 & 100.0 & 96.3 & 58.6 & 59.2 & 100.0 & 99.0 & 77.5 & 73.2 \\
transistor & 100.0 & 99.3 & 92.6 & 85.7 & 93.7 & 98.0 & 86.7 & 79.6 & 99.5 & 98.0 & 87.3 & 81.9 & 100.0 & 99.9 & 98.2 & 93.2 & 100.0 & 99.6 & 95.2 & 89.5 \\
wood & 99.4 & 98.9 & 84.6 & 74.5 & 99.9 & 99.4 & 91.6 & 83.8 & 99.6 & 99.0 & 87.0 & 79.6 & 99.8 & 99.4 & 89.4 & 81.1 & 100.0 & 99.2 & 90.3 & 82.1 \\
zipper & 100.0 & 99.6 & 86.0 & 79.2 & 100.0 & 99.6 & 90.7 & 82.7 & 100.0 & 99.7 & 88.2 & 81.6 & 100.0 & 99.5 & 88.5 & 82.0 & 100.0 & 99.3 & 88.3 & 79.9 \\
\midrule
Average & \underline{99.7} & 99.1 & 82.0 & 77.1 & 98.9 & 99.1 & 84.5 & 78.7 & 99.6 & 98.7 & 83.1 & 78.1 & \textbf{99.8} & \underline{99.2} & \underline{86.3} & \underline{80.8} & \textbf{99.8} & \textbf{99.3} & \textbf{88.1} & \textbf{82.2} \\
\bottomrule
\end{tabular}
\caption{MVTec AD category-level results. Bold/underline denote the best/second-best Average scores.}
\label{tab:mvtec_main}
\end{table*}

\section{Experiments}

\subsection{Experimental Setup}

\textbf{Protocol.}
We evaluate on MVTec AD~\cite{mvtecad} and VisA~\cite{visa}
following AnomalyDiffusion~\cite{hu2023anomalydiffusion}: one third of
each anomaly type is used for generator adaptation and the remaining
two thirds for testing. For each anomaly type, every method generates
1000 synthetic image--mask pairs to train the same downstream anomaly
inspection model with identical scripts, which is evaluated on real
test anomalies. The complete dataset split, training
hyperparameters, score aggregation, and metric definitions are provided
in Appendix~A of the supplementary material.

\textbf{Metrics.}
We report AP-I, AUC-P, AP-P, and F1-P, equally averaged across
categories. Auxiliary quality is measured by KID~\cite{kid},
IC-LPIPS~\cite{iclpips}, and the classification accuracy (Cls Acc) of
a ResNet-34~\cite{resnet} trained on each method's synthetic images.

\textbf{Baselines.}
We compare \method{} with AnomalyDiffusion~\cite{hu2023anomalydiffusion},
DualAnoDiff~\cite{dualanodiff}, SeaS~\cite{dai2025SeaS}, and
O2MAG~\cite{rao2026one}, using official implementations and recommended
settings whenever available. DefectFill~\cite{song2025defectfill} is
excluded because its evaluation applies generation to masks from the
held-out two-thirds split, violating our protocol that prohibits test
anomaly masks. Among the compared methods, O2MAG is the most recent and
strongest-performing baseline and therefore serves as our primary point
of comparison. Since O2MAG lacks an independent mask generator, its
full-pair setting combines its images with AnomalyDiffusion-style
generated masks following its reported practice, without using real test
masks.

\begin{figure}[t]
\centering
\includegraphics[width=0.95\columnwidth]{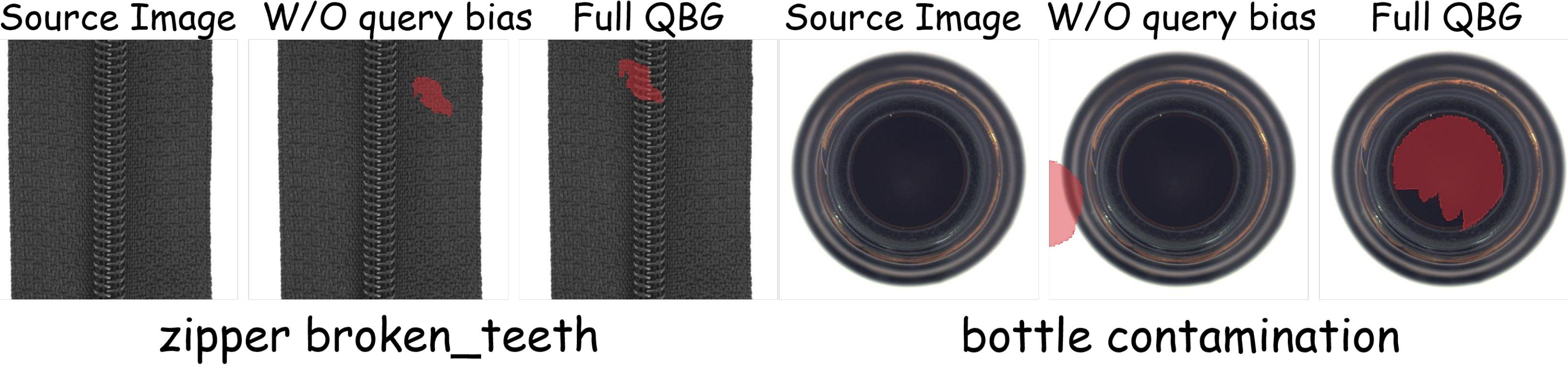}
\caption{Qualitative comparison of mask placement with and without query
bias. Full \QBG{} improves object compatibility; red overlays denote
generated masks.}
\label{fig:qbg_compare}
\end{figure}

\subsection{Main Results}

\begin{table}[t]
\centering
\small
\begin{tabular}{lcccc}
\toprule
Method & AP-I & AUC-P & AP-P & F1-P \\
\midrule
AnomalyDiffusion & 88.9 & 97.0 & 42.9 & 45.8 \\
DualAnoDiff & \underline{95.4} & 98.7 & 61.3 & 62.1 \\
SeaS & 93.3 & \underline{98.8} & 65.1 & 64.1 \\
O2MAG & 94.0 & \textbf{98.9} & \underline{67.2} & \underline{64.6} \\
\method{} & \textbf{96.6} & 98.2 & \textbf{68.5} & \textbf{66.1} \\
\bottomrule
\end{tabular}
\caption{Anomaly localization on VisA, averaged over 12 categories. Bold/underline denote best/second-best results.}
\label{tab:visa_main}
\end{table}

\begin{table}[t]
\centering
\small
\begin{tabular}{lccc}
\toprule
Method & KID$\downarrow$ & IC-LPIPS$\uparrow$ & Cls Acc$\uparrow$ \\
\midrule
AnomalyDiffusion & 102.67 & 0.29 & 70.25 \\
DualAnoDiff & 103.87 & \textbf{0.38} & 65.55 \\
SeaS & 125.67 & 0.35 & 56.70 \\
O2MAG & 45.55 & 0.30 & 82.35 \\
\method{} & \textbf{39.70} & 0.29 & \textbf{82.51} \\
\bottomrule
\end{tabular}
\caption{Generation-quality and classification-based auxiliary evaluation on MVTec AD.}
\label{tab:generation_quality}
\end{table}

\begin{figure}[t]
\centering
\includegraphics[width=0.95\columnwidth]{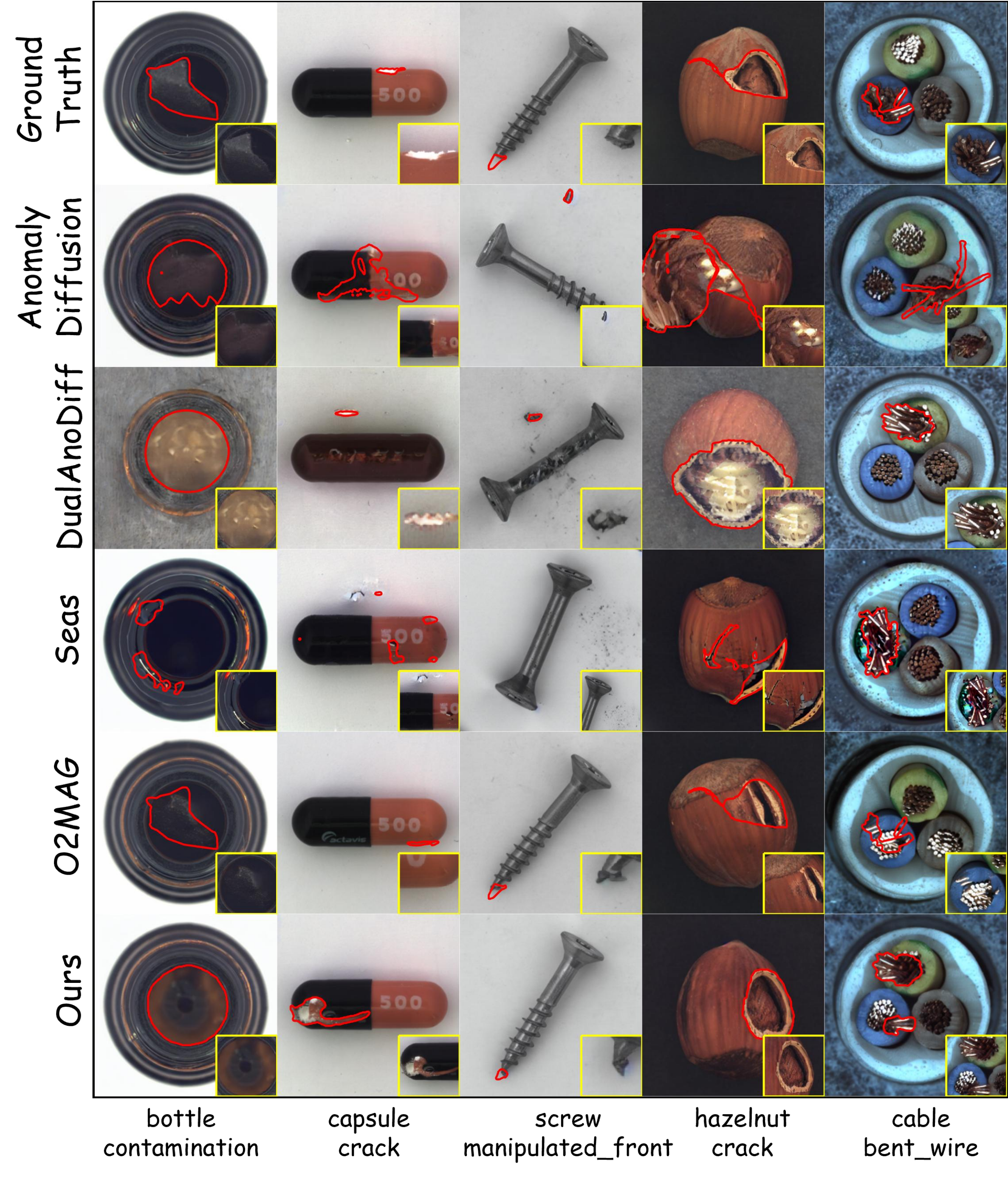}
\caption{Qualitative comparison on representative MVTec AD categories.
Red contours denote anomaly masks, and yellow boxes show enlarged
regions. \method{} improves mask alignment, defect realism, and
background preservation.}
\label{fig:qualitative}
\end{figure}

Tables~\ref{tab:mvtec_main} and~\ref{tab:visa_main} show that
\method{} achieves the best AP-P/F1-P on MVTec AD and VisA,
reaching 88.1/82.2 and 68.5/66.1, respectively, while remaining
competitive in AP-I and AUC-P. Table~\ref{tab:generation_quality}
further reports the lowest KID and highest Cls Acc, indicating a
distribution closer to real anomalies with identifiable defect
semantics. We treat IC-LPIPS only as an auxiliary indicator because
background corruption can also inflate it~\cite{choi2026magic,rao2026one}.
Qualitative examples are shown in Figure~\ref{fig:qualitative}.

\subsection{Ablation Study}
We evaluate the components on six categories: cable, capsule, grid,
and screw from MVTec AD, and fryum and pcb3 from VisA. As shown in
Table~\ref{tab:ablation}, removing \QBG{} lowers AP-P by 8.6 points;
Figure~\ref{fig:qbg_compare} further shows that query bias improves
object-compatible mask placement. Under the same 8k-step budget,
removing Stages~1--3 reduces AP-P by
2.1, 5.4, and 1.5 points, respectively, with Stage~2 contributing
most among the three stages. Disabling \ISC{} lowers AP-P/F1-P by
4.6/3.8 points, confirming complementary inference-time gains. The
lower block validates $\mathcal{L}_{\mathrm{attn}}$ and
$\mathcal{L}_{\mathrm{bg}}$; clean-latent locking with $\hat{x}_0$
outperforms the noise-space background constraint.

\begin{table}[t]
\centering
\small
\begin{tabular}{lcccc}
\toprule
Configuration & AP-I & AUC-P & AP-P & F1-P \\
\midrule
Full \method{} & 98.9 & 97.3 & 65.7 & 62.8 \\
\midrule
\multicolumn{5}{l}{\emph{Core components}}\\
w/o Query-Bias-Gen & 98.0 & 96.8 & 57.1 & 57.5 \\
w/o Stage 1 & 98.9 & 97.2 & 63.6 & 60.6 \\
w/o Stage 2 & 98.9 & 95.4 & 60.3 & 58.4 \\
w/o Stage 3 & 98.5 & 97.1 & 64.2 & 61.8 \\
w/o \ISC{} & 98.7 & 94.9 & 61.1 & 59.0 \\
\midrule
\multicolumn{5}{l}{\emph{Staged training objective}}\\
w/o $\mathcal{L}_{\mathrm{attn}}$ & 98.8 & 96.9 & 63.5 & 60.9 \\
w/o $\mathcal{L}_{\mathrm{bg}}$ & 98.9 & 97.0 & 62.4 & 60.6 \\
$\mathcal{L}_{\mathrm{bg}}$ on noise & 98.6 & 97.3 & 62.2 & 59.6 \\
\bottomrule
\end{tabular}
\caption{Ablation results averaged over six categories. Stage-removal
variants use the same 8k-step training budget. Top: core components;
bottom: staged-training losses.}
\label{tab:ablation}
\end{table}

\subsection{Diagnostics and Efficiency}

\paragraph{Label-source diagnostic.}
To disentangle image quality from label construction, we evaluate
O2MAG and \method{} on the six-category ablation subset using
identical normal sources and augmented condition masks. For each generator, we fix the generated images and use either the
augmented masks or labels recovered by the same materialization
predictor under the same protocol. Thus, between-generator comparisons
isolate image quality, while within-generator comparisons quantify
materialization.

\begin{table}[t]
\centering
\small
\setlength{\tabcolsep}{2.5pt}
\begin{tabular*}{0.856\columnwidth}{@{\extracolsep{\fill}}llcccc@{}}
\toprule
Generator & Label source & AP-I & AUC-P & AP-P & F1-P \\
\midrule
O2MAG
& Aug. mask
& 93.9 & 95.4 & 41.2 & 44.5 \\
O2MAG
& Materialized
& 95.9 & 96.1 & 54.1 & 55.3 \\
\midrule
\method{}
& Aug. mask
& 97.9 & 96.5 & 59.3 & 58.8 \\
\method{}
& Materialized
& \textbf{98.9} & \textbf{97.3} & \textbf{65.7} & \textbf{62.8} \\
\bottomrule
\end{tabular*}
\caption{Controlled label-source diagnostic on six categories.
Both generators use identical sources and augmented masks;
``Aug. mask'' uses them directly, while ``Materialized'' uses
labels recovered by the same predictor.}
\label{tab:label_source}
\end{table}

Table~\ref{tab:label_source} shows that \method{} outperforms O2MAG
by 18.1/14.3 AP-P/F1-P points with augmented-mask labels and by
11.6/7.5 points after materialization, showing that the gain does not
arise solely from label recovery. Materialization improves O2MAG by 12.9/10.8 AP-P/F1-P points and
\method{} by 6.4/4.0, indicating a general label-alignment benefit;
the smaller gain for \method{} suggests better intrinsic alignment
between its visible defects and spatial support.

\paragraph{Mask over-following diagnostic.}
To test mask over-following, we compare the full model with
the w/o Stage~2 variant under equal training budgets and identical
source--mask--seed triplets, using enlarged masks derived from real
anomalies. AR-FR measures altered pixels inside the condition region,
whereas S-FR extracts semantically visible anomalies through
materialization. Removing Stage~2 increases the mean AR-FR from
45.5 to 47.9 and S-FR from 66.0 to 69.8, indicating greater anomaly
spread over the condition region. Category-level results, threshold
robustness, and the complete protocol are provided in Appendix~I.

\paragraph{Object-awareness diagnostic.}

We fix the mask generator and compare matched-source conditioning with a
same-category shuffled source. For each object--anomaly category, normal
training images and their SAM3~\cite{sam3} foregrounds are jointly
rotated by $\{0^\circ,\pm45^\circ,\pm90^\circ\}$, yielding 500 masks per
anomaly category.

We quantify mask--object compatibility using the Object Hit Rate
(OHR), defined as the fraction of the generated mask area that falls
inside the target object foreground:
\begin{equation}
    \mathrm{OHR}(M,R)=\frac{|M\cap R|}{|M|},
    \label{eq:ohr}
\end{equation}
where $M$ and $R$ are the generated mask and target foreground.
We report OHR at $0^\circ$ and its mean over the four nonzero rotations
(OHR@rot), equally aggregated across samples and categories. Their
difference is computed as
$\Delta=\mathrm{OHR}@0^\circ-\mathrm{OHR}@\mathrm{rot}$.
The complete source-pairing protocol, rotation construction, sample
budget, empty-mask handling, and aggregation order are provided in
Appendix~J of the supplementary material.

\begin{figure}[t]
\centering
\includegraphics[width=0.95\columnwidth]{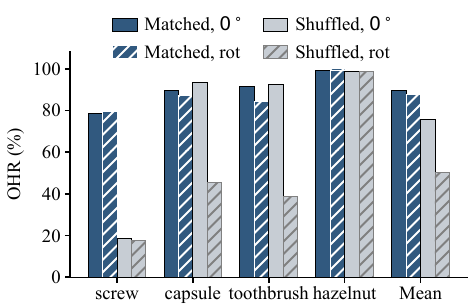}
\caption{Object-awareness diagnostic on four MVTec AD categories.
``rot'' denotes the mean OHR over four nonzero rotations. Matched and
shuffled sources use the target image and another same-category image,
respectively.}
\label{fig:ohr_diagnostic}
\end{figure}

Matched-source OHR remains $87.5\%$ under rotation, compared with
$50.1\%$ for shuffled sources, corresponding to drops of $2.1$ and
$25.7$ points, respectively (Figure~\ref{fig:ohr_diagnostic}).
Directional categories show the strongest contrast, while the
rotationally symmetric hazelnut acts as a control, confirming
source-structure-aware mask placement.

\paragraph{Cost analysis.}
On one NVIDIA A100, \method{} incurs a one-time cost of $2.39$ hours per
category and $4.65$ seconds per generated pair.
It is cheaper than AnomalyDiffusion, DualAnoDiff, and O2MAG in both one-time and per-pair costs. Only SeaS is cheaper, but
its AP-P is $5.0$ points lower. Full accounting is provided in Appendix~L.

\section{Conclusion}

We present \method{}, an object-aware image--mask generation framework
for localization, addressing instance-incompatible masks and the coupling
between defect learning and spatial control. \QBG{} conditions mask
diffusion on a matched normal image to produce object-aware priors, while
the three-stage curriculum learns normal appearance, defect semantics
under loose conditions, and fine-grained mask calibration. During
sampling, \ISC{} restricts anomaly propagation and preserves normal
content, while materialization recovers pixel-level labels aligned with
realized defects. Experiments on MVTec AD and VisA show that \method{}
produces synthetic pairs and balances localization, generation quality,
and computational efficiency. Future work will reduce category-specific
adaptation costs and improve generalization to unseen objects and
industrial scenarios.

\bibliography{anogdraftrefs}

\end{document}